\documentclass[journal]{IEEEtran}
\usepackage{amsmath,amsfonts}
\usepackage{amsthm,amssymb}
\usepackage{mathrsfs}

\usepackage{algorithmic}
\usepackage{algorithm}
\usepackage{array}
\usepackage[caption=false,font=normalsize,labelfont=sf,textfont=sf]{subfig}
\usepackage{textcomp}
\usepackage{stfloats}
\usepackage{url}
\usepackage{verbatim}
\usepackage{graphicx}
\usepackage{cite}

\usepackage{soul}
\usepackage{bm}
\usepackage{booktabs}
\usepackage{txfonts}
\usepackage{booktabs}
\usepackage{multirow}
\usepackage{caption3}
\usepackage{xcolor}
\usepackage[table]{xcolor}
\usepackage{makecell}
\usepackage{float}      
\usepackage{adjustbox}  

\usepackage{listings}
\usepackage{pifont}
\usepackage{threeparttable}

\begin{document}
\title{HiSem: Hierarchical Semantic Disentangling for Remote Sensing Image Change Captioning
}

\author{Man Wang$^{1,\dagger}$, Chenyang Liu$^{1,\dagger}$, Wenjun Li, Feng Ni, Bing Jia, Baoqi Huang, Riting Xia$^*$, \\and Zhenwei Shi$^*$,~\IEEEmembership{Senior Member,~IEEE}

\thanks{The work was supported by the National Natural Science Foundation of China under Grant 62125102 and 624B2017, 62506177, the Inner Mongolia Autonomous Region Science and Technology Planning Project under Grants 2025YFHH0124, the fund of supporting the reform and development of local universities (Disciplinary construction), and the fund of First-class Discipline Special Research Project of Inner Mongolia A. R. of China under Grant YLXKZXND-036. 
( \emph{$^{\dagger}$:Equal contribution; 
$^{*}$:Corresponding authors.})
}

\thanks{Man Wang, Bing Jia, Baoqi Huang, Riting Xia and Zhenwei Shi are with the College of Computer Science, Inner Mongolia University, Hohhot 010021, China, and with the Research Center for Spatiotemporal Intelligence, Inner Mongolia University, Hohhot 010021, China (e-mail: shizhenwei@imu.edu.cn; xiart@imu.edu.cn).

Chenyang Liu and Zhenwei Shi are with the Department of Aerospace Intelligent Science and Technology, School of Astronautics, Beihang University, Beijing 100191, China, and also with the State Key Laboratory of Virtual Reality Technology and Systems, Beihang University, Beijing 100191, China.

Wenjun Li and Feng Ni are with the China Mobile Communications Group Inner Mongolia Co., Ltd. Hohhot, Inner Mongolia Autonomous Region 010010, China.

Chenyang Liu is also with Shen Yuan Honors College of Beihang University, Beijing 100191, China.
}
}

\maketitle

\begin{abstract}
Remote sensing image change captioning (RSICC) aims to achieve high-level semantic understanding of genuine changes occurring between bi-temporal images. Despite notable progress, existing methods are fundamentally limited by a shared modeling assumption: changed and unchanged image pairs, which have intrinsically different semantic granularities, are processed under a unified modeling strategy. This modeling inconsistency leads to semantic entanglement between coarse-grained change existence judgment and fine-grained semantic understanding.
To address the above limitation, we propose a novel hierarchical semantic disentangling network (HiSem) that explicitly disentangles semantic representations of different granularities. 
Specifically, we first introduce the Bidirectional Differential Attention Modulation (BDAM) module that leverages discrepancy-aware attention to enhance cross-temporal interactions, thereby amplifying true change signals while suppressing irrelevant variations. 
Building upon this, we design a Hierarchical Adaptive Semantic Disentanglement (HASD) module that performs adaptive routing at two hierarchical levels: a coarse-grained image-level routing mechanism distinguishes changed and unchanged image pairs, while a fine-grained token-level Mixture-of-Experts (MoE) block models diverse and heterogeneous change semantics for changed samples.
Extensive experiments on two benchmark datasets demonstrate that HiSem outperfoms previous methods, achieving a significant improvement of +7.52\% BLEU-4 on the WHU-CDC dataset. More importantly, our approach provides a structured perspective for RSICC by explicitly aligning model design with the intrinsic semantic heterogeneity of bi-temporal scenes.
The code will be available at \emph{\url{https://github.com/Man-Wang-star/HiSem}}

\end{abstract}

\begin{IEEEkeywords}
Remote sensing, change captioning, hierarchical semantic disentanglement, transformer.
\end{IEEEkeywords}

\section{Introduction}
\IEEEPARstart{R}{emote} sensing image change captioning (RSICC) is an emerging vision-language task in the remote sensing community. Given a pair of remote sensing images (RSIs) capturing the same geographical area at different times, the RSICC task aims to automatically describe the differences between bi-temporal images in natural language~\cite{chang2023changes,liu2023decoupling}. RSICC technology plays an important role in various fields, such as land planning~\cite{Liu_2022MLAT, albarakati2024novel,chen2025dynamicvis}, urbanization analysis~\cite{liu2020building,chen2025rsrefseg}, and environmental monitoring~\cite{malila1980change,liu2024changeAgent}.

RSICC can be viewed as a natural evolution of remote sensing change analysis toward richer semantic understanding. Early binary change detection (BCD) methods mainly focus on pixel-level localization, answering where changes occur while largely ignoring their semantic meaning~\cite{changeformer,zhang2024bifa,CDMamba}. Semantic change detection (SCD) further introduces category-level supervision to identify what type of objects have changed, yet it still remains primarily a visual classification task limited to predefined semantic labels~\cite{ding2024joint,guo2025taco,chen2023TTP}. RSICC pushes this paradigm one step further by introducing natural language as the output modality. Instead of simply identifying change regions or categories, RSICC aims to generate human-readable descriptions that explicitly explain what has changed and how these changes manifest over time, including object attributes, spatial relationships, and temporal evolution~\cite{zhou2024single,RSICCformer,sun2024lightweight}.

\begin{figure}
	\centering
	\includegraphics[width=1\linewidth]{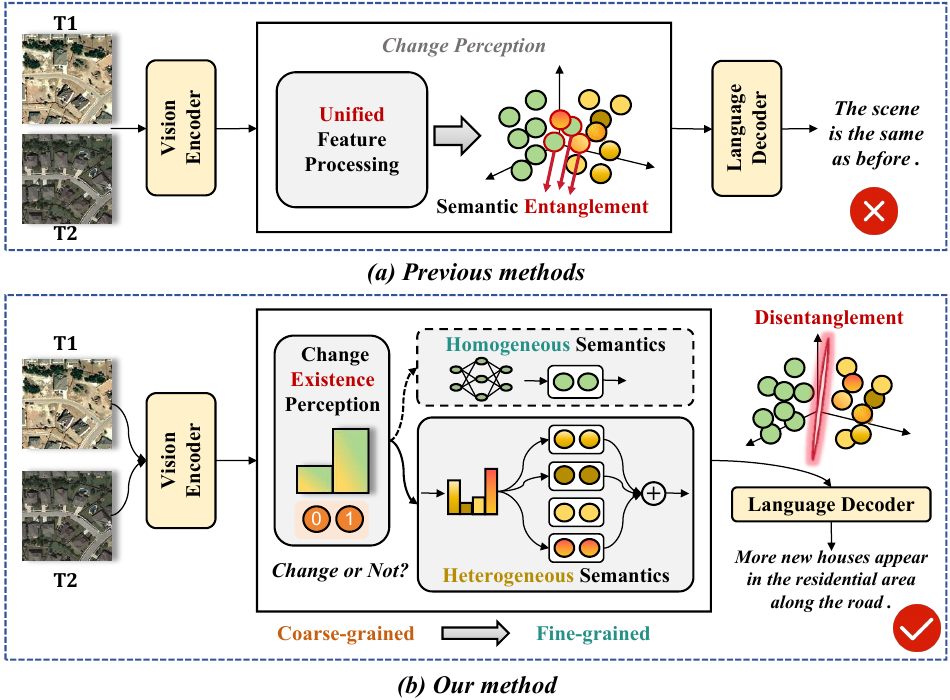}
	\caption{(a) Previous methods uniformly process changed and unchanged image pairs, resulting in semantic entanglement. (b) Our method explicitly distinguishes between coarse-grained existence perception and fine-grained semantic understanding.
}
	\label{fig:1}
\end{figure}

Most existing RSICC methods are built upon an encoder-decoder framework and generally follow a three-stage paradigm: bi-temporal visual representation, change perception, and language generation~\cite{zhou2024single,RSICCformer,sun2024lightweight,liu2024MV_CC,li2025cd4c,KCFI}, as illustrated in Fig.~\ref{fig:1}(a). In the visual representation stage, Siamese pretrained backbone networks, such as Convolutional Neural Networks (CNNs)~\cite{Resnet} or Vision Transformers (ViTs)~\cite{Vit}, are commonly employed to extract spatially structured and semantically rich features from bi-temporal images. In the change perception stage, dedicated interaction mechanisms are introduced to explicitly model cross-temporal relationships between paired features, enabling the model to highlight semantically meaningful differences while suppressing irrelevant variations. Finally, language models (e.g, Transformer~\cite{Transformer}, LSTM~\cite{LSTM}, and Mamba~\cite{gu2023mamba}) map the resulting change-aware visual representations into coherent natural language.

Within this framework, the quality of change-aware representations learned during the change perception stage largely determines the accuracy of the generated captions. However, remote sensing temporal scenes are inherently complex and often contain numerous irrelevant variations caused by illumination differences, seasonal effects, or sensor noise. These factors introduce two intertwined challenges for RSICC: 
1) suppressing irrelevant variations to accurately determine whether changes exist and where they occur, and 
2) extracting fine-grained semantic information from true change regions to support coherent and informative language generation.
To address these challenges, most existing studies focus on improving the change perception stage. For example, Li et al.~\cite{li2024inter} proposed a cross-temporal attention (CTA) mechanism to model direct interactions between concatenated bi-temporal features, strengthening salient change cues while suppressing background interference. 
To further enrich semantic modeling, Sun et al.~\cite{sun2025scene} constructed a scene graph based on difference features and employed a Graph Convolutional Network (GCN) to capture object attributes and inter-object relationships within change regions.

Despite these advances, existing methods still suffer from a fundamental limitation. Most approaches adopt a unified modeling strategy for both changed and unchanged image pairs, implicitly assuming that they can be processed at the same semantic granularity. In practice, however, unchanged image pairs usually correspond to simple descriptions (e.g., ``No change has occurred"), whereas changed image pairs require modeling diverse and fine-grained semantics (e.g., ``A building is constructed adjacent to the road"). Treating these two fundamentally different cases in a uniform modeling manner leads to \textbf{semantic entanglement between coarse-grained change existence judgment and fine-grained change semantic understanding}. 
This observation motivates the need for a more structured change-aware modeling approach that explicitly disentangles semantic representations across different granularities.

To address the above limitation, we propose a hierarchical semantic disentangling network (HiSem) that explicitly disentangles semantic representations of different granularities, as illustrated in Fig.~\ref{fig:1} (b). Specifically, we first employ a Vision Transformer of the pretrained CLIP model~\cite{radford2021learning} to extract semantic representations for bi-temporal images. To obtain more discriminative bi-temporal features, we introduce a Bidirectional Differential Attention Modulation (BDAM) module, which leverages bi-temporal difference features as attention bias to explicitly modulate cross-temporal feature interactions in both “before-to-after” and “after-to-before” directions. This design enhances bi-temporal features of genuinely changed regions while suppressing irrelevant temporal variations.

Furthermore, to achieve semantic disentanglement between changed and unchanged image pairs, we propose a Hierarchical Adaptive Semantic Disentanglement (HASD) module that separately models their semantics. At the coarse-grained level of change existence perception, an image-level routing mechanism is employed to determine whether an image pair is changed or unchanged based on enhanced difference features, and to adaptively select corresponding processing paths. At the fine-grained semantic modeling level, changed image pairs, which exhibit complex and diverse semantic changes, are processed by a Mixture-of-Experts (MoE) based block, where each expert specializes in capturing distinct implicit change semantics. In contrast, unchanged image pairs, whose semantics are relatively homogeneous and simple, are routed through a lightweight block to extract coarse-grained invariant representations, thereby preserving computational efficiency. Finally, the disentangled visual representations are fed into a Transformer-based decoder to generate natural language descriptions that faithfully reflect the temporal semantic states between bi-temporal images.

By aligning model design with the intrinsic semantic heterogeneity of bi-temporal scenes, HiSem provides a structured solution that bridges coarse-grained change existence perception and fine-grained semantic understanding, enabling more accurate and interpretable change captioning.

Our contributions are summarized as follows:

\begin{itemize}
\item 
Unlike previous methods that uniformly process changed and unchanged image pairs, resulting in semantic entanglement, we propose a hierarchical semantic disentangling model that distinguishes between coarse-grained change existence perception and fine-grained semantic understanding to alleviate the semantic entanglement.

\item 
To achieve semantic disentanglement, the BDAM module leverages bi-temporal difference features as attention bias to guide cross-temporal feature interactions and enhance discriminative change cues, while the HASD module performs coarse-to-fine semantic routing to adaptively model changed and unchanged samples with different semantic granularities.

\item 
Extensive experiments on two large benchmark datasets demonstrate that HiSem consistently outperforms previous methods with a significant margin (+7.52\% BLEU-4 on WHU-CDC), validating the effectiveness of the proposed hierarchical semantic disentanglement approach.

\end{itemize}

\section{Related work}
Remote sensing image change captioning (RSICC) has emerged as an important task in the intelligent interpretation of remote sensing imagery \cite{liu2025remote, weng2025vision, chen2025rscc,MetaEarth,yuan2025autodrive,yuan2025video,yuan2026if} and has attracted increasing research attention in recent years. Most existing methods follow a three-stage paradigm: (1) bi-temporal visual representation, (2) change perception, and (3) language generation.

{\textbf{1) Bi-temporal Visual Representation.}}
In the first stage, Siamese or dual-stream architectures with CNNs or Vision Transformers (ViTs) pretrained on large-scale natural image datasets are commonly used to extract generic spatial and semantic features from bi-temporal images~\cite{RSICCformer,chang2023changes,cai2023interactive}.
Considering the inherent scale variability in remote sensing imagery, multi-scale feature extraction is further introduced to capture changes at different spatial resolutions, where methods based on ResNet101 \cite{he2016deep, sun2024lightweight, vyshnav2024intelli} exploit hierarchical CNN features, while Transformer-based encoders such as SegFormer \cite{xie2021segformer, wu2025cross, zou2025frequency,chen2023rsprompter} provide multi-scale representations with enhanced global context modeling.

To mitigate the domain gap between natural image pretraining and downstream bi-temporal remote sensing imagery, Zhou \textit{et al.} \cite{zhou2024single} proposed a single-stream extractor network (SEN) that applies contrastive learning to concatenated bi-temporal RSIs, aiming to improve feature transferability while maintaining computational efficiency. Furthermore, Yang \textit{et al.} \cite{yang2024remote} introduced a hierarchical multi-attentive network (MADiffCC) integrated with a diffusion model, which is pretrained on a remote sensing image dataset to effectively model data distributions and temporal-spatial contextual information.

{\textbf{2) Change Perception.}}
The change perception stage aims to learn discriminative representations for actual changes while suppressing irrelevant variations such as illumination differences, seasonal effects, or sensor noise. From a methodological perspective, prevailing research focuses on optimizing bi-temporal feature fusion, which can be systematically categorized into spatial-level, temporal-level and joint spatio-temporal fusion.

Early research primarily emphasizes spatial-level fusion, modeling discrepancies between bi-temporal feature maps to achieve precise change localization. RSICCformer \cite{RSICCformer} bolsters change-aware representations by utilizing cross-temporal feature differences as keys to modulate original features. 
Diverging from this explicit difference-guided approach, Chg2Cap \cite{chang2023changes} adopts an alternative strategy that first refines individual temporal embeddings to capture intra-information, subsequently integrating them via concatenation and stacked attention layers to model inter-temporal correlations. 
SCNet~\cite{sun2026scnet} further highlights the importance of channel-wise dependencies, introducing a spatial-channel attention encoder to sequentially capture inter-region relationships and semantic attributes. 
MAF-Net \cite{chen2024multi} aggregates paired hierarchical bitemporal features through a Multi-scale Change Aware Encoder, utilizing Transformer-based difference and content cross-attention to capture discriminative change information across diverse scales.
While effective at capturing spatial cues, these approaches remain limited in modeling temporal evolution.

To address temporal dependencies, MFRNet \cite{xu2024mfrnet} adopts a joint attention mechanism that utilizes concatenated bi-temporal features as queries to mutually guide attention distributions across temporal branches. Li \textit{et al.} \cite{li2024intertemporal} enhance temporal modeling by explicitly constraining the consistency between “before-to-after” and “after-to-before” transitions. This alignment of symmetrical change features is achieved via a bi-directional triplet ranking loss, which is incorporated to enforce symmetry in the latent space.
Despite these advances, many methods still neglect the intrinsic correlations between spatial and temporal cues, yielding incoherent spatio-temporal representations. RSCaMa~\cite{liu2024rscama} addresses this by leveraging the Mamba architecture and State Space Models (SSMs) to jointly model spatial textures and temporal dynamics within a unified representation. Recognizing that spatio-temporal features alone remain sensitive to environmental variations, FST-Net \cite{rs17081463} further broadens the scope of joint modeling by incorporating a frequency-domain decomposition mechanism.

From the perspective of auxiliary information integration, several studies leverage change detection to enhance change perception in RSICC.
For instance, Pix4Cap \cite{liu2024pixel} incorporates an auxiliary change detection branch supervised by pseudo-labels from the pre-trained change detection model to provide spatial constraints for change localization. Furthermore, Semantic-CC \cite{zhu2024semantic} incorporates foundational semantic priors from the Segment Anything Model (SAM) \cite{kirillov2023segment} via a multi-task semantic aggregation neck to mediate the interaction between change detection and RSICC features, facilitating pixel-level semantic guidance for sentence generation. Whereas Li \textit{et al.} \cite{li2024detection} directly leverage P2V-CD \cite{lin2022transition} model to generate explicit change masks, which are element-wise multiplied with the original images to suppress task-irrelevant background noise while preserving critical spatial details.

Despite these advances, most methods still process changed and unchanged image pairs at a uniform semantic granularity, leading to potential semantic entanglement between coarse-grained change existence and fine-grained change semantics. To address this limitation, our HiSem network explicitly disentangles semantic representations across two levels: coarse-grained change identification via the BDAM module, which guides cross-temporal interactions using spatial priors, and fine-grained modeling of heterogeneous change patterns through the HASD module, which leverages MoE-based experts to adaptively capture diverse change semantics.

{\textbf{3) Language Generation.}}
In the language generation stage, language models (e.g, LSTM~\cite{LSTM}, Transformer~\cite{Transformer}, and Mamba~\cite{gu2023mamba}) map the change-aware visual representations into coherent natural language descriptions. Early methods primarily relied on Recurrent Neural Networks (RNNs) or Support Vector Machines (SVMs) for sequential description generation \cite{chouaf2021captioning, hoxha2022change}. 
With the rapid development of deep learning, Transformer-based language decoders have become the predominant choice \cite{RSICCformer,chang2023changes,Peng2024GRSL_RSICC,cai2023interactive}, owing to their strong capability in modeling cross-modal dependencies via cross-attention mechanisms. 
To further improve the quality of generated descriptions, several studies focus on improving vision-language attention mechanisms to enable more effective cross-modal interaction\cite{yang2024remote,cai2023interactive,liu2023progressive}. For example, MADiffCC \cite{yang2024remote} introduces a gated multi-head cross-attention (GMCA) mechanism to adaptively select and aggregate essential multi-scale features for enhanced change description generation.
However, the quadratic complexity of Transformers limits their scalability. Accordingly, some studies explore more efficient State Space Models (SSMs) \cite{gu2023mamba,chen2024rsmamba,qinzhe2026state}-based decoders with linear complexity. A representative example is Liu et al. \cite{liu2024rscama}, which explores the Mamba-based decoder for efficient caption generation.

More recently, large language models (LLMs) have been introduced to RSICC to improve semantic understanding and caption generation. Existing studies can be broadly categorized into two lines according to their adaptation strategies: prompt or instruction-based adaptation and parameter-efficient fine-tuning (PEFT).
Prompt-based methods aim to adapt LLMs without updating their parameters. Prompt-CC \cite{liu2023decoupling} introduces a multi-prompt learning strategy to condition pre-trained LLMs on image-level and class-specific cues, enabling effective change-aware caption generation via conditional prompting. KCFI \cite{KCFI} utilizes key change features as visual instructions into align visual and linguistic modalities, enabling the frozen LLMs to generate more accurate change-aware descriptions through task-specific fine-tuning.
Furthermore, parameter-efficient fine-tuning approaches have been explored to balance performance and efficiency, including low-rank adaptation (LoRA) based methods \cite{hu2022lora, deng2025changechat, irvin2024teochat, elgendy2024geollava,liu2025text2earth} and adapter-based methods \cite{wang2024chareption, wang2024ccexpert}. These approaches introduce a small number of trainable parameters while keeping the original parameters of LLMs frozen, enabling efficient adaptation to RSICC tasks. However, such over-parameterized LLMs often impose substantial computational overhead.

\begin{figure*}
	\centering
    \includegraphics[width=1.0\linewidth]{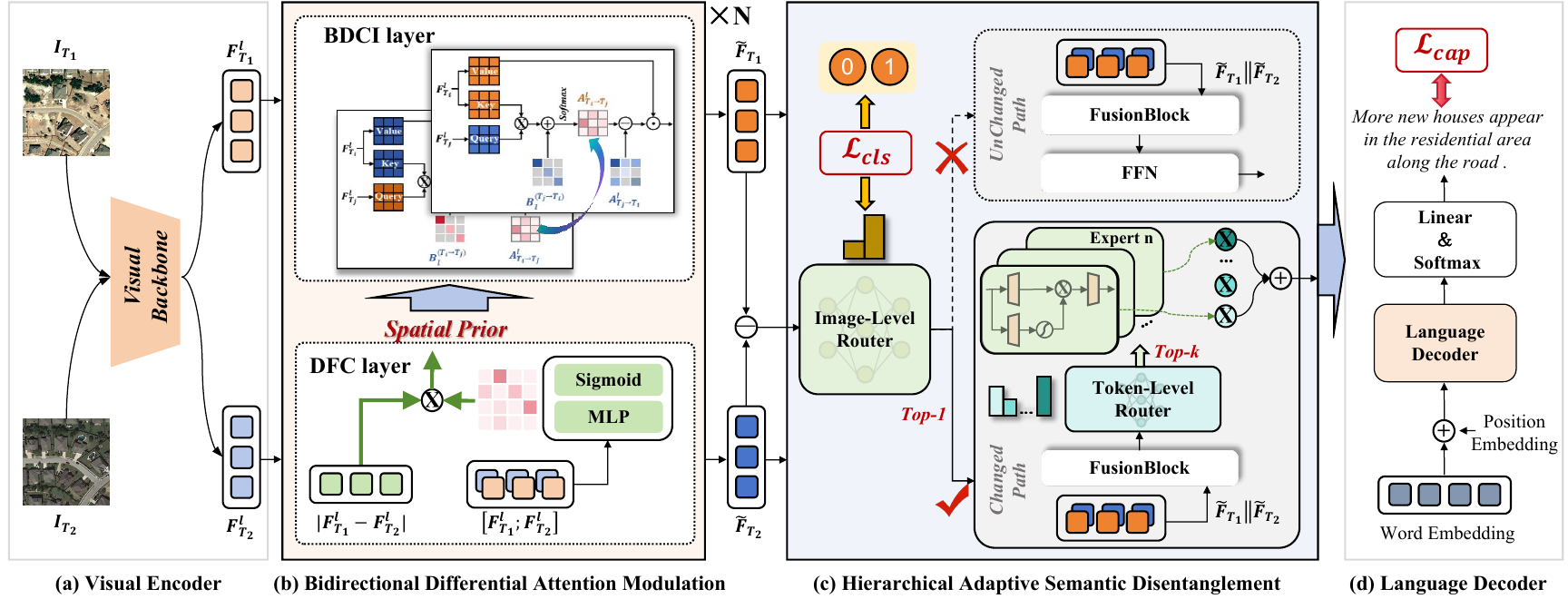}
	\caption{Overview of our proposed HiSem, which consists of four components: Visual Encoder, Bidirectional Differential Attention Modulation (BDAM), Hierarchical Adaptive Semantic Disentanglement (HASD), and Language Decoder.
 }
	\label{fig:2}
\end{figure*}

\section{Methodology}

\subsection{Overview of HiSem}
As illustrated in Fig.~\ref{fig:2}, the proposed HiSem network consists of four core components: a visual encoder, a Bidirectional Differential Attention Modulation (BDAM) module, a Hierarchical Adaptive Semantic Disentanglement (HASD) module, and a language decoder. Given a bi-temporal image pair $(I_{T_1}, I_{T_2})$, we first employ a CLIP-based \cite{radford2021learning} vision encoder pretrained on large-scale image-text data to extract semantically rich representations. The resulting visual embeddings are denoted as $\mathbf{F}_{T_1}, \mathbf{F}_{T_2} \in \mathbb{R}^{B \times L \times D}$, where $B$ is the batch size, $L$ is the number of visual tokens, and $D$ is the feature dimension. Positional embeddings are added to preserve spatial structure prior to further processing.

The BDAM module then refines these visual embeddings to enhance change-sensitive regions. It comprises two sequential sublayers per layer: (i) a discrepancy-aware feature conditioning sublayer that highlights semantically informative difference channels, and (ii) a bidirectional discrepancy-guided cross-temporal interaction sublayer that explicitly modulates cross-temporal feature interactions in both ``before-to-after'' and ``after-to-before'' directions using learnable difference priors. This design strengthens genuine change features while suppressing irrelevant temporal variations.

Following BDAM, the HASD module performs hierarchical semantic disentanglement. At the coarse level, an image-level routing mechanism classifies pairs as changed or unchanged, directing them into corresponding expert paths. Unchanged pairs are processed by a lightweight block to extract invariant semantics efficiently, whereas changed pairs are routed to a MoE-based block to capture diverse and fine-grained change semantics. Finally, the disentangled visual representations are fed into a Transformer-based language decoder, generating natural language descriptions that faithfully reflect temporal semantic information between bi-temporal images.

\subsection{Bidirectional Differential Attention Modulation}
\begin{figure}
	\centering
	\includegraphics[width=1.0\linewidth]{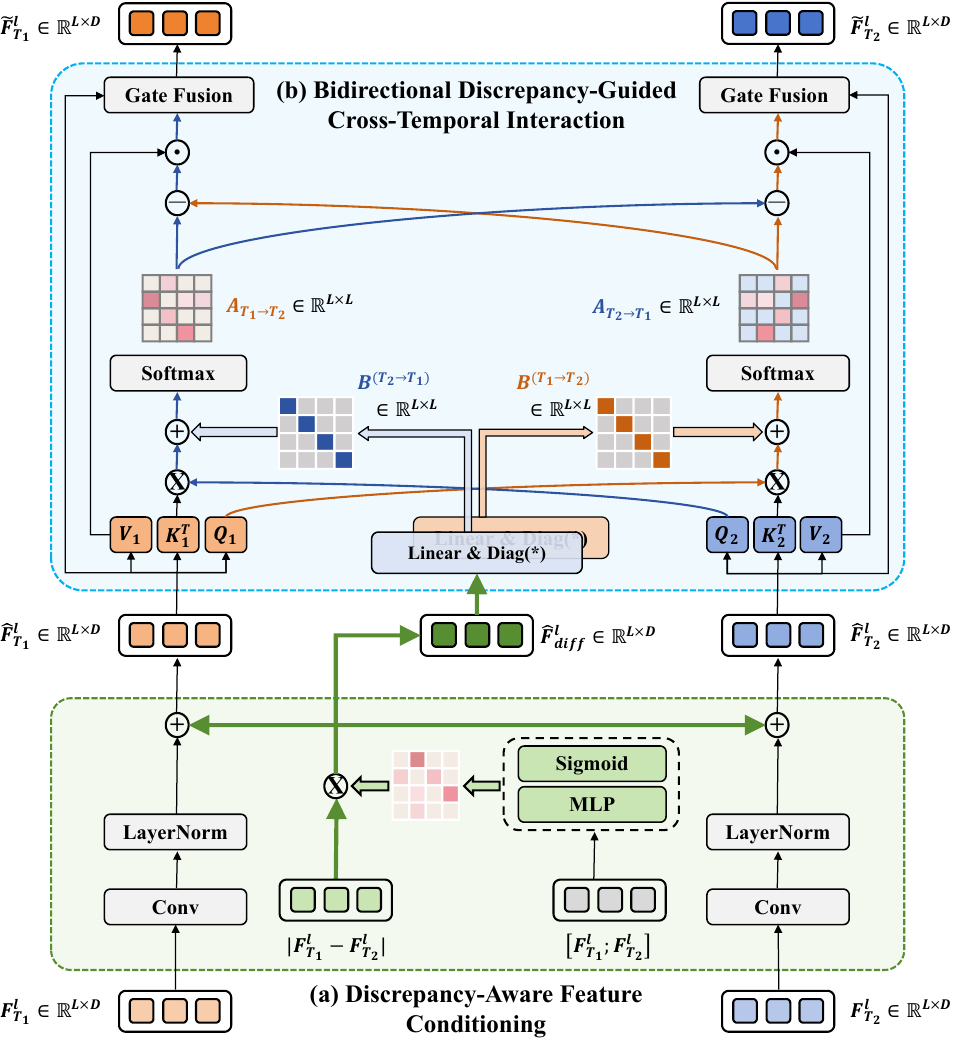}
	\caption{The structure of the BDAM. (a) Discrepancy-aware feature conditioning sublayer reweights the aggregated bi-temporal features to spatially enhance change-sensitive responses. (b) Bidirectional discrepancy-guided cross-temporal interaction sublayer exploits the difference features as learnable attention cues to explicitly modulate cross-temporal interactions, capturing temporal dependencies between bi-temporal images.
 }
	\label{fig:3}
\end{figure}

As illustrated in Fig.~\ref{fig:3}, the proposed Bidirectional Differential Attention Modulation (BDAM) module is designed to progressively refine bi-temporal representations by explicitly incorporating difference-driven cues.
Each BDAM layer follows a spatial-to-temporal progressive modulation strategy, consisting of two sequential sublayers: (i) discrepancy-aware feature conditioning for spatially localized change enhancement, and (ii) bidirectional discrepancy-guided cross-temporal interaction for modeling directional change dependencies.

\textbf{1) Discrepancy-Aware Feature Conditioning Sublayer}:
The first sublayer aims to enhance change-sensitive responses while suppressing semantically irrelevant variations caused by illumination differences, seasonal effects, or sensor noise.
Given bi-temporal embeddings $\mathbf{F}_{T_1}^l, \mathbf{F}_{T_2}^l \in \mathbb{R}^{L \times D}$ at the $l$-th BDAM layer, we first compute absolute difference features:
\begin{equation}
\mathbf{F}_{\text{diff}}^l = \lvert \mathbf{F}_{T_1}^l - \mathbf{F}_{T_2}^l \rvert. 
\end{equation}

Rather than directly using these raw differences, we introduce a lightweight conditioning mechanism to adaptively reweight feature channels according to their relevance to change perception.
Specifically, a two-layer MLP generates channel-wise modulation weights from concatenated bi-temporal features:
\begin{align}
\mathbf{G}_{\text{cond}}^l &= \sigma \Big( \mathbf{W}_2  \phi \big( \mathbf{W}_1 [ \mathbf{F}_{T_1}^l ; \mathbf{F}_{T_2}^l ] \big) \Big), \\
\mathbf{\hat{F}}_{\text{diff}}^l &= \mathbf{F}_{\text{diff}}^l \odot \mathbf{G}_{\text{cond}}^l,
\end{align}
where $[\cdot;\cdot]$ denotes channel-wise concatenation, $\phi(\cdot)$ and $\sigma(\cdot)$ represent ReLU and sigmoid activations, respectively, and $\odot$ denotes element-wise multiplication.

To preserve the local spatial structure, each temporal feature map is reshaped into its 2D form and processed using convolution. The resulting features are then conditioned by the reweighted difference representations $\mathbf{\hat{F}}_{\text{diff}}^l$, yielding spatially enhanced bi-temporal features $\mathbf{\hat{F}}_{T_1}^l$ and $\mathbf{\hat{F}}_{T_2}^l$.
This sublayer provides stable and localized discrepancy cues, which serve as a foundation for subsequent cross-temporal interaction.

\textbf{2) Bidirectional Discrepancy-Guided Cross-Temporal Interaction Sublayer}:
While the previous sublayer enhances local discrepancy cues, effective change modeling further requires capturing temporal dependencies between bi-temporal features. To this end, we introduce a bidirectional discrepancy-guided cross-temporal interaction mechanism. Specifically, we first leverage the discrepancy-enhanced features $\mathbf{\hat{F}}_{\text{diff}}^l$ to construct a structured, position-wise attention bias:
\begin{equation}
\mathbf{B}^{(d)} = \alpha^{(d)} \cdot \mathrm{{Diag}} \big( \mathbf{W}^{(d)} \mathbf{\hat{F}}_{\text{diff}}^l \big), \quad d \in \{ T_1 \rightarrow T_2, \; T_2 \rightarrow T_1 \},
\end{equation}
where $\mathbf{W}^{(d)}$ is a learnable projection that aggregates channel-wise discrepancy responses into scalar positional scores, $\alpha^{(d)}$ is a learnable scaling factor, and $\mathrm{{Diag}}(\cdot)$ denotes a diagonal matrix.
This diagonal bias emphasizes intrinsic change intensity at each spatial location and acts as a structured modulation term for subsequent attention logits.

Then, bidirectional cross-temporal attention is computed by alternately using each temporal feature as the query and the other as key and value. The bias $\mathbf{B}^{(d)}$ is added to the attention logits in each direction to emphasize spatial locations with strong temporal discrepancies. Formally:
\begin{align}
\mathbf{A}_{T_1 \rightarrow T_2} &= \mathrm{Softmax}\Big( \frac{\mathbf{Q}_{T_1} \mathbf{K}_{T_2}^\top}{\sqrt{d}} + \mathbf{B}^{(T_1\rightarrow T_2)} \Big), \\
\mathbf{A}_{T_2 \rightarrow T_1} &= \mathrm{Softmax}\Big( \frac{\mathbf{Q}_{T_2} \mathbf{K}_{T_1}^\top}{\sqrt{d}} + \mathbf{B}^{(T_2\rightarrow T_1)} \Big),
\end{align}
where $\mathbf{Q}_{T_i}$ and $\mathbf{K}_{T_i}$ are the projected queries and keys, $d$ is the key dimension, and $\mathbf{B}^{(d)}$ explicitly biases attention toward positions with higher direction-aware temporal discrepancies.

To further emphasize direction-aware temporal discrepancies, we combine the two directional attention responses via a differential operation:
\begin{align}
\mathbf{F}^O_{T_1 \rightarrow T_2} &= \big( \mathbf{A}_{T_1 \rightarrow T_2} - \lambda \mathbf{A}_{T_2 \rightarrow T_1} \big) \mathbf{V}_{T_2}, \\
\mathbf{F}^O_{T_2 \rightarrow T_1} &= \big( \mathbf{A}_{T_2 \rightarrow T_1} - \lambda \mathbf{A}_{T_1 \rightarrow T_2} \big) \mathbf{V}_{T_1},
\end{align}
where $\mathbf{V}_{T_i}$ denotes the projection of the visual feature and $\lambda$ is a balancing coefficient.
By contrasting the two directional attentions, invariant responses are attenuated while asymmetric discrepancy cues are highlighted, without discarding contextual information entirely.

Finally, the resulting features are adaptively fused with the original bi-temporal representations through a gated mechanism, yielding temporally enhanced features $\mathbf{\tilde{F}}_{T_1}^l$ and $\mathbf{\tilde{F}}_{T_2}^l$.
These representations provide robust, discrepancy-aware change cues and are subsequently fed into the hierarchical adaptive semantic disentanglement module for coarse-to-fine semantic modeling.

\subsection{Hierarchical Adaptive Semantic Disentanglement}
\begin{figure}
	\centering
	\includegraphics[width=1.0\linewidth]{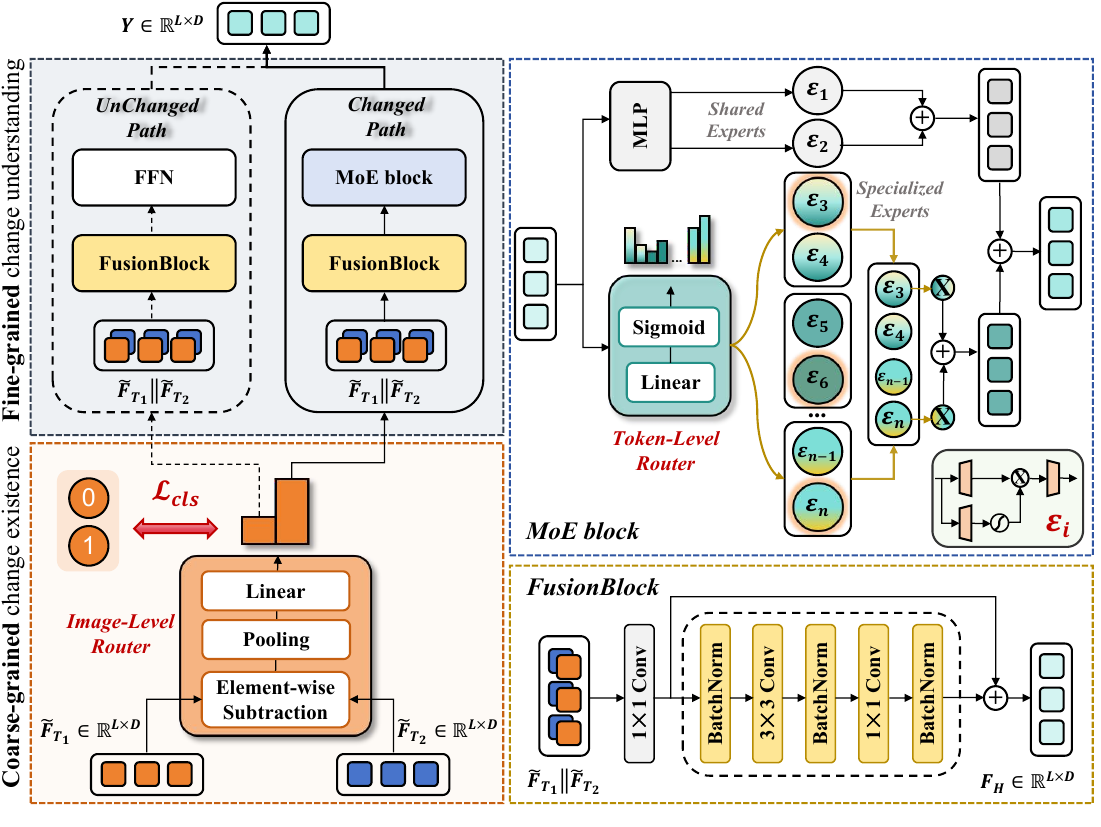}
	\caption{The proposed hierarchical semantic disentanglement mechanism provides a structured perspective for RSICC by decomposing the task into two sequential stages: coarse-grained change existence perception driven by an image-level router and fine-grained change semantic understanding driven by a token-level router. 
 }
	\label{fig:4}
\end{figure}

To achieve comprehensive and fine-grained semantic understanding of bi-temporal RSIs, we propose a hierarchical adaptive semantic disentanglement network that explicitly separates the modeling of changed and unchanged image pairs, as illustrated in Fig.~\ref{fig:4}. This design follows a coarse-to-fine strategy: first, coarse-grained image-level routing determines the existence of changes, and then fine-grained token-level routing determines the activated expert groups to captures diverse change semantics.

\textbf{1) Coarse-Grained Image-Level Routing:}
At the coarse level, we design a difference-driven image-level routing mechanism to explicitly model the existence of changes in bi-temporal image pairs. 

Given the enhanced features $\mathbf{\tilde{{F}}}_{T_1}$ and $\mathbf{\tilde{{F}}}_{T_2}$ from BDAM, we first compute their difference and apply global average pooling (GAP) to obtain a compact image-level representation:
\begin{align}
\mathbf{\hat{Z}} &= \mathrm{GAP}\big(\mathbf{\tilde{{F}}}_{T_2} - \mathbf{\tilde{{F}}}_{T_1}\big) \mathbf{W}_g,
\end{align}
where $\mathbf{W}_g \in \mathbb{R}^{D \times E}$ is a learnable routing matrix mapping the $D$-dimensional feature to $E$ expert paths (in our case, $E=2$ corresponding to changed and unchanged paths).

The routing logits $\mathbf{\hat{Z}}$ are then converted to probabilities via softmax, and the most relevant expert path is selected using a top-1 routing strategy:
\begin{align}
\left( \mathcal{E}_{k}, \boldsymbol{\alpha}_{k} \right)
&= \operatorname*{Top\text{-}k}\!\left(\mathbf{\mathrm{Softmax\left({\mathbf{\hat{Z}}}\right)}}\right),
\qquad\sum_{e \in \mathcal{E}_{k}} \alpha_{e} = 1,
\end{align}
where $\mathcal{E}_k$ denotes the selected processing path index and $\boldsymbol{\alpha}_k$ contains the corresponding normalized routing weights. Specifically, $\mathcal{E}_k = 1$ indicates assignment to the changed-path expert, while $\mathcal{E}_k = 0$ corresponds to the unchanged-path expert.

To ensure semantically consistent and stable routing decisions, the predicted routing path is supervised with a cross-entropy classification loss:
\begin{align}
\label{cls_loss}
\mathcal{L}_{\mathrm{cls}} = \mathrm{CE}(\mathbf{\hat{Z}}, Z),
\end{align}
where $Z \in {0,1}$ denotes the ground-truth change label for the image pair. This coarse-grained routing mechanism effectively integrates change existence judgment into the network, providing a reliable foundation for the subsequent fine-grained token-level semantic disentanglement.

\textbf{2) Fine-Grained Token-Level Disentanglement:}
Once the expert path is determined by the coarse-grained image-level routing, the enhanced bi-temporal features are first aggregated via a lightweight FusionBlock equipped with convolutional operations to preserve spatio-temporal contextual information:
\begin{align}
\mathbf{F}_H &= \mathrm{FusionBlock}\!\left(\tilde{\mathbf{F}}_{T_1} \,\Vert\,\tilde{\mathbf{F}}_{T_2}\right),
\end{align}
where $\Vert$ denotes concatenation along the feature dimension.

For \emph{unchanged} image pairs, the fused representations mainly encode temporally invariant semantics. Such representations can be effectively modeled using a simple feed-forward network (FFN).

In contrast, for \emph{changed} image pairs, semantic changes are often complex and highly heterogeneous across tokens. A single transformation is therefore insufficient to model these diverse change patterns. To address this challenge, we introduce a MoE-based block that enables adaptive, token-wise semantic modeling, allowing different experts to specialize in distinct types of fine-grained change semantics. Given the fused features $\mathbf{F}_H$, we perform token-level routing to dynamically assign each token to a small subset of specialized experts. Specifically, a lightweight gating module computes expert scores for each token:
\begin{align}
\mathbf{S}_{\text{expert}}
&=\sigma\!\left(\mathbf{F}_H\mathbf{W}_{c}\right),
\end{align}
where $\mathbf{S}_{\text{expert}} \in \mathbb{R}^{L \times E}$ denotes the expert scores for all $L$ tokens over $E$ experts, $\sigma(\cdot)$ is the sigmoid activation, and $\mathbf{W}_c \in \mathbb{R}^{D \times E}$ is a learnable routing matrix. In our design, $E=8$ specialized experts are employed.

To mitigate uneven expert activation and encourage balanced utilization of specialized experts, we adopt a group-constrained routing \cite{liu2024deepseek} strategy. Specifically, expert relevance scores are first aggregated at the group level by applying max pooling over experts within each group:
\begin{align}
\mathbf{S}_{\text{group}}
&=\operatorname{MaxPool}_{\text{group}}\bigl(\mathbf{S}_{\text{expert}}\bigr), 
\end{align}
where each group score $\mathbf{S}_{\text{group}}$ reflects the strongest expert response in that group, serving as a proxy for the group’s relevance to the current token.

Based on the resulting group-level scores, we select the top-$k_g$ most relevant expert groups for each token, thereby constraining the subsequent expert selection to a compact and semantically coherent subset of experts:
\begin{align}
\mathbf{\mathcal{G}}_{k_g}(n)
&=
\operatorname*{Top\text{-}k_g}
\bigl(
\mathbf{S}_{\text{group}}(n,:)
\bigr). \label{eq:group_topk} 
\end{align}

Then, expert-level sparse routing is conducted conditioned on the group-level selection $\mathbf{\mathcal{G}}_{k_g}(n)$. For each token $n$, routing is confined to experts within the selected groups, producing the group-filtered expert scores $\tilde{\mathbf{S}}_{\text{expert}}$. The top-$k$ experts are subsequently activated according to their routing scores:
\begin{align}
\bigl(\mathcal{E}_k(n), \boldsymbol{\alpha}_k(n)\bigr) &= \operatorname*{Top\text{-}k}\bigl(\tilde{\mathbf{S}}_{\text{expert}}(n,:)\bigr),
\quad \sum_{e \in \mathcal{E}_k(n)} \alpha_e = 1.
\end{align}

The outputs of the activated specialized experts are aggregated via a weighted sum according to their routing scores $\alpha_e$, ensuring that each token integrates information from the most relevant experts. This aggregated representation is then combined with the output of the shared experts $f_{\text{s}}(\cdot)$, which capture global semantic information common to all tokens, yielding the final token representation:
\begin{align}
\mathbf{Y} &= \sum_{e \in \mathcal{E}_k} \alpha_e f_e(\mathbf{F}_H) + f_{\text{s}}(\mathbf{F}_H),
\end{align}
where $f_i(\cdot)$ denotes a specialized expert implemented with a $\text{SwiGLU}$ \cite{shazeer2020glu} feed-forward network, with $i \in \{e, s\}$.

\subsection{Multi-Stage Curriculum Learning Strategy}
The proposed model is trained in a fully supervised manner, where the primary objective is bi-temporal change captioning. Given a predicted word distribution and the corresponding ground-truth caption, we optimize the model using the standard cross-entropy loss:
\begin{equation}
\mathcal{L}_{\mathrm{cap}} = - \frac{1}{N} \sum_{n=1}^{N} \sum_{v=1}^{V} y_{n}^{(v)} \log p_{n}^{(v)},
\end{equation}
where $N$ is the total number of words, $V$ is the vocabulary size, $y_{n}^{(v)}$ denotes the one-hot ground-truth indicator at position $n$, and 
$p_{n}^{(v)}$ is the predicted probability of the $v$-th word.

In addition to caption generation, we introduce an auxiliary image-level classification objective (Eq.~\ref{cls_loss}) to supervise the coarse-grained routing in HASD. This auxiliary task serves a supportive role by enforcing semantically consistent expert-path selection, rather than acting as an independent prediction objective. However, we observe that directly jointly optimizing the captioning loss and the routing classification loss from scratch often leads to unstable training performance.

To address this issue, we adopt a multi-stage curriculum learning strategy that gradually introduces auxiliary supervision. Specifically, during the warm-up stage, gradients from the image-level routing branch are blocked by detaching the routing classification logits. This allows the model to first focus on learning robust visual-linguistic alignment and stable caption generation without interference from premature routing supervision. After the warm-up stage, the auxiliary routing classification loss is progressively activated using a cosine-based ramp-up schedule over the remaining epochs. This smooth transition prevents abrupt optimization shifts and enables the routing mechanism to adapt in accordance with increasingly reliable visual and linguistic representations. The overall training objective is formulated as:
\begin{align}
\mathcal{L} = \mathcal{L}_{\text{cap}} + \lambda\, \alpha(e)\, \mathcal{L}_{\text{cls}},
\end{align}
where $\mathcal{L}_{\mathrm{cls}}$ denotes the cross-entropy loss for image-level change classification, $\lambda$ is a balancing coefficient set to $0.8$, and $\alpha(e)$ is an epoch-dependent cosine ramp-up factor.

The proposed multi-stage curriculum learning strategy effectively stabilizes training while ensuring that the auxiliary routing supervision is introduced in a controlled and semantically meaningful manner, ultimately leading to more reliable hierarchical semantic disentanglement and improved captioning performance.

\section{Experiments}
\subsection{Datasets}
We evaluate our model HiSem on two widely used large benchmarksdatasets LEVIR-CC~\cite{RSICCformer} and WHU-CDC \cite{shi2024multi} for RSICC.
LEVIR-CC~\cite{RSICCformer} is a large-scale benchmark for RSICC, containing 10,077 bi-temporal image pairs (5,038 exhibiting semantic changes and 5,039 unchanged) annotated with 50,385 human-annotated captions. The maximum caption length is 39 words, with an average sentence length of 7.99 words. All images are collected from 20 geographically distinct regions across Texas, USA, with a temporal range spanning from 5 to 15 years. Each image is $256 \times 256$ pixels in size with a spatial resolution of 0.5 m/pixel. 

WHU-CDC \cite{shi2024multi} comprises 7,434 high-resolution bi-temporal image pairs, annotated with a total of 37,170 descriptive captions. Each image is $256 \times 256$ pixels in size with a spatial resolution of 0.075 m/pixel. The dataset maintains a balanced ratio of changed and unchanged pairs.
Caption annotations range from 3 to 24 words in length, with a total vocabulary of 327 unique words.

\subsection{Evaluation Metrics}
To rigorously assess the quality of the generated captions, we employ four widely used evaluation metrics, including BLEU-N (N=1,2,3,4), ROUGE$_L$, METEOR, and CIDEr-D. 
BLEU-N evaluates the precision of generated captions by comparing their $n$-grams with reference sentences, whereas ROUGE$_L$ focuses on recall by measuring the longest common subsequences between predictions and references. METEOR integrates both precision and recall while accounting for synonyms and stemming variations. CIDEr-D evaluates both the accuracy and diversity of generated captions by comparing their word frequency patterns with those of multiple reference captions, emphasizing consistency with the consensus across all references. Higher values on these metrics reflect closer agreement with the reference captions, indicating superior quality of the generated sentences. Besides, following previous methods, we adopt an average metric $S^*_m$ as follows:
\begin{equation}
S^*_m = \frac{1}{4}\mathrm{(BLEU\text{-}4+ROUGE_L+METEOR+CIDEr\text{-}D)}.
\end{equation}

\subsection{Implementation Details}
All models are implemented in PyTorch and trained on an NVIDIA GTX 4090 GPU.
CLIP-ViT-B/16 \cite{radford2021learning} is adopted as the visual encoder, while a Transformer-based decoder is employed for language generation. Model optimization is performed using the Adam optimizer with an initial learning rate of $1\times10^{-4}$. The model is trained for 50 epochs with a batch size of 64. During inference, beam search is applied with a beam size of 1.
The proposed BDAM module comprises three stacked layers, whereas the HASD module contains a single layer. The Transformer decoder is configured with one layer, and the embedding dimension is set to 768. To ensure statistical reliability, all experimental results are reported as the average over five independent runs.
\begin{table*}
\renewcommand{\arraystretch}{1.3}
\caption{Performance comparison of our proposed method with previous state-of-the-art methods on the LEVIR-CC dataset. The bolded results are the best.
}
\label{tab:Comparisons_other_methods_LEVIR-CC}
\centering
\resizebox{0.95\linewidth}{!}{
\begin{tabular}{c | c c c c c c c | c}
	\toprule
	Method & BLEU-1 & BLEU-2 & BLEU-3 & BLEU-4 & METEOR & ROUGE$_L$ & CIDEr-D & $S^*_m$\\
	\midrule
    {RSICCFormer\cite{RSICCformer} {\textcolor{gray}{[TGRS'22]}}} & 84.72 & 76.27 & 68.87 & 62.77 & 39.61 & 74.12 & 134.12 & 77.66 \\
    {Chg2Cap\cite{chang2023changes} {\textcolor{gray}{[TIP'23]}}} & 86.14 & 78.08 & 70.66 & 64.39 & 40.03 & 75.12 & 136.61 & 79.04 \\
    {PSNet\cite{liu2023progressive} {\textcolor{gray}{[IGARSS'23]}}} & 83.86 & 75.13 & 67.89 & 62.11 & 38.80 & 73.60 & 132.62 & 76.78 \\ 
    {CTMTNet\cite{shi2024multi} {\textcolor{gray}{[TGRS'24]}}} & 85.95 & 77.99 & 70.74 & 64.69 & 39.49 & 74.54 & 134.94 & 78.42 \\
    {Sen\cite{zhou2024single} {\textcolor{gray}{[TGRS'24]}}} & 85.10 & 77.05 & 70.01 & 64.09 & 39.59 & 74.57 & 136.02 & 78.57 \\
    {SFT\cite{sun2024lightweight} {\textcolor{gray}{[JSTARS'24]}}} & 84.56 & 75.87 & 68.64 & 62.87 & 39.93 & 74.69 & 137.05 & 78.64 \\
    {Pix4Cap\cite{liu2024pixel} {\textcolor{gray}{[IGARSS'24]}}} & 85.56 & 77.08 & 69.79 & 63.78 & 39.96 & 75.12 & 136.76 & 78.91 \\
    {RSCaMa\cite{liu2024rscama} {\textcolor{gray}{[GRSL'24]}}} & 85.79 & 77.99 & 71.04 & 65.24 & 39.91 & 75.24 & 136.56 & 79.24 \\ 
    {Diffusion-RSCC\cite{yu2025diffusion} {\textcolor{gray}{[TGRS'25]}}} & - & - & - & 60.90 & 37.80 & 71.50 & 125.60 & 73.95 \\
    {Mask Approx Net\cite{sun2025mask} {\textcolor{gray}{[TGRS'25]}}} & 85.90 & 77.12 & 70.72 & 64.32 & 39.91 & 75.67 & 137.71 & 79.40 \\
    {Prompt-CC\cite{liu2023decoupling} {\textcolor{gray}{[TGRS'23]}}} & 83.66 & 75.73 & 69.10 & 63.54 & 38.82 & 73.72 & 136.44 & 78.13 \\
    {Chareption\cite{wang2024chareption} {\textcolor{gray}{[PRCV'24]}}} & 83.82 & 75.39 & 68.37 & 62.53 & 40.38 & 74.72 & 137.83 & 77.58 \\
	{KCFI\cite{KCFI} {\textcolor{gray}{[TIP'25]}}} & 86.34 & 77.31 & 70.89 & 65.30 & 39.42 & 75.47 & 138.25 & 79.61 \\
    \rowcolor{gray!15}
\textbf{HiSem (Ours)}  
 & \textbf{86.60} & \textbf{78.74} & \textbf{71.73} & \textbf{65.82} & \textbf{40.39} & \textbf{75.77}	& \textbf{138.86} & \textbf{80.21} \\
	\bottomrule
\end{tabular}
}
\end{table*}

\begin{table*}[ht] 
\renewcommand{\arraystretch}{1.3}
\caption{Performance comparison of our proposed method with previous state-of-the-art methods on the WHU-CDC dataset. The bolded results are the best.
}
\label{tab:Comparisons_other_methods_WHU-CDC}
\centering
\resizebox{0.95\linewidth}{!}{
\begin{tabular}{c | c c c c c c c | c}
	\toprule
	Method & BLEU-1 & BLEU-2 & BLEU-3 & BLEU-4 & METEOR & ROUGE$_L$ & CIDEr-D & $S^*_m$\\
	\midrule
    {RSICCFormer\cite{RSICCformer} {\textcolor{gray}{[TGRS'22]}}} & 80.05 & 74.24 & 69.61 & 66.54 & 42.65 & 73.91 & 133.44 & 79.14 \\
    {Chg2Cap\cite{chang2023changes} {\textcolor{gray}{[TIP'23]}}} & 78.93 & 72.64 & 67.20 & 62.71 & 41.46 & 77.95 & 144.18 & 81.58 \\
    {PSNet\cite{liu2023progressive} {\textcolor{gray}{[IGARSS'23]}}} & 81.26 & 73.25 & 65.78 & 60.32 & 36.97 & 71.60 & 130.52 & 74.85 \\ 
    {Sen\cite{zhou2024single} {\textcolor{gray}{[TGRS'24]}}} & 80.60 & 74.64 & 67.69 & 61.97 & 36.76 & 71.70 & 133.57 & 76.00 \\
    {SFT\cite{sun2024lightweight} {\textcolor{gray}{[JSTARS'24]}}} & 81.17 & 72.90 & 66.06 & 60.27 & 37.34 & 72.63 & 134.64 & 76.22 \\
    {Prompt-CC\cite{liu2023decoupling} {\textcolor{gray}{[TGRS'23]}}} & 81.12 & 73.96 & 67.22 & 61.45 & 36.99 & 71.88 & 134.50 & 76.21 \\
    {Diffusion-RSCC\cite{yu2025diffusion} {\textcolor{gray}{[TGRS'25]}}} & 75.32 & 70.15 & 66.40 & 63.76 & 40.18 & 73.80 & 127.96 & 76.43 \\
    {Mask Approx Net\cite{sun2025mask} {\textcolor{gray}{[TGRS'25]}}} & 81.34 & 75.68 & 71.16 & 67.73 & 43.89 & 75.41 & 135.31 & 80.59 \\
    {CTMTNet\cite{shi2024multi} {\textcolor{gray}{[TGRS'24]}}} & 83.56 & 77.66 & 72.76 & 69.00 & 45.39 & 79.23 & 149.40 & 85.76 \\
    \rowcolor{gray!15}
\textbf{HiSem (Ours)}  
 & \textbf{87.49} & \textbf{82.89} & \textbf{79.17} & \textbf{76.52} & \textbf{48.77} & \textbf{82.00} & \textbf{158.35} & \textbf{91.41} \\
	\bottomrule
\end{tabular}
}
\end{table*}

\begin{figure*}
	\centering
    \includegraphics[width=1.0\linewidth]{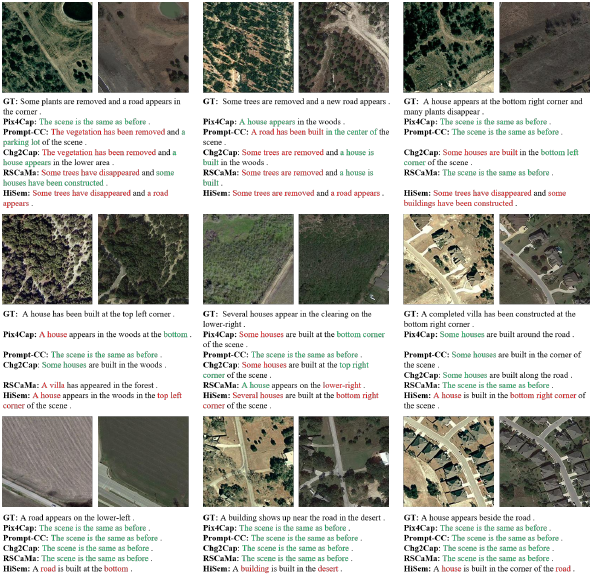}
	\caption{Captions generated by Pix4Cap, Prompt-CC, Chg2Cap, RSCaMa, and our HiSem on the LEVIR-CC dataset. \textbf{GT} denotes one of the five ground-truth captions. Text highlighted in \textcolor{red}{red} corresponds to more accurate and detailed descriptions, whereas text highlighted in \textcolor{green}{green} represents incorrect descriptions.
 }
	\label{fig:caption}
\end{figure*}

\subsection{Comparison to State-of-the-Art}
We compare the proposed method with several representative state-of-the-art RSICC approaches on the LEVIR-CC and WHU-CDC benchmarks. To provide a more structured comparison, we organize the selected baselines according to their decoder architectures. They are grouped into three categories: Transformer-based decoders, LLM-based decoders, and emerging alternatives.
Transformer-based decoders represent the dominant paradigm in RSICC, including RSICCFormer \cite{RSICCformer}, PSNet \cite{liu2023progressive}, CTMTNet \cite{shi2024multi}, Sen \cite{zhou2024single}, SFT \cite{sun2024lightweight}, Pix4Cap \cite{liu2024pixel}, Chg2Cap \cite{chang2023changes}, and Mask Approx Net \cite{sun2025mask}. These methods generally employ cross-attention-based visual-language interaction to decode change-aware visual representations into textual descriptions.
LLM-based decoders include Prompt-CC \cite{liu2023decoupling}, Chareption \cite{wang2024chareption}, and KCFI \cite{KCFI}, which leverage pretrained large language models to enhance caption generation by aligning visual change features with rich linguistic priors.
Beyond these two primary categories, several recent studies explore emerging alternatives. Diffusion-RSCC \cite{yu2025diffusion} formulates caption generation as a progressive denoising process rather than conventional autoregressive decoding, while RSCaMa \cite{liu2024rscama} adopts a Mamba-based sequence modeling framework as an efficient alternative to Transformer-based decoders for long-range context modeling.
Despite these differences in decoder design, the compared methods generally follow an encoder-decoder framework, in which the encoder primarily aims to perceive and encode discriminative visual changes from bi-temporal images, and the decoder subsequently maps these change-aware representations to natural language.

As reported in Tables \ref{tab:Comparisons_other_methods_LEVIR-CC} and \ref{tab:Comparisons_other_methods_WHU-CDC}, HiSem consistently achieves state-of-the-art performance across all evaluation metrics on both datasets. On the LEVIR-CC dataset, HiSem surpasses the previous best-performing method KCFI with consistent gains. More notably, on the WHU-CDC dataset, HiSem achieves substantial improvements over previous methods $(+7.52\%$ on BLEU-4, $+3.38\%$ on METEOR, $+2.77\%$ on ROUGE$_L$, $+8.95\%$ on CIDEr-D, and $+5.65\%$ on $S^*_m)$, indicating stronger capability in generating detailed and semantically faithful change descriptions.
These performance gains reveal a key limitation shared by many existing approaches: they implicitly treat changed and unchanged samples at a uniform semantic granularity, assuming that all bi-temporal image pairs require equally rich semantic modeling. In contrast, our method explicitly decouples the semantic representations of these two categories, enabling differentiated processing. The proposed BDAM module introduces a change-aware bidirectional differential attention mechanism that suppresses irrelevant background responses while selectively amplifying discriminative change cues. Building upon this, the HASD module further enables adaptive semantic routing by explicitly disentangling changed and unchanged samples into distinct processing paths. Through hierarchical semantic modeling, MoE-based experts can focus on fine-grained change semantics without interference from unchanged samples, while maintaining compact and invariant representations for stable scenes. This design leads to more semantically consistent and informative captions, as evidenced by the pronounced gains in the evaluation metrics, particularly on WHU-CDC.

Overall, these results demonstrate that our proposed hierarchical semantic disentanglement idea provides a more effective modeling perspective for RSICC, especially in scenarios involving complex and heterogeneous change distributions, validating the robustness and generalization capability of the proposed HiSem method.

\subsection{Qualitative Results}
\subsubsection{\textbf{Qualitative Comparison Results}}
Fig.~\ref{fig:caption} presents qualitative comparisons of captioning results between the proposed HiSem method and several state-of-the-art approaches, including Pix4Cap~\cite{liu2024pixel}, Prompt-CC~\cite{liu2023decoupling}, Chg2Cap~\cite{chang2023changes}, and RSCaMa~\cite{liu2024rscama}, on the LEVIR-CC dataset. In the figure, red highlights indicate more accurate and detailed descriptions, whereas green highlights denote incorrect descriptions. 

In the first row, the four image pairs involve both appearance and disappearance change patterns. Our proposed HiSem accurately captures both change states within the same pair, whereas other methods tend to miss one of the changes or produce inaccurate descriptions. 
The second row features complex scenarios with multiple objects and inter-object spatial relationships. Our HiSem generates comprehensive and fine-grained descriptions, correctly reflecting object locations, relationships, and multiple changed objects.
For instance, in the first image pair of the second row, our method generates a detailed description of locations (e.g., ``A house appears in the woods in the \textbf{top left corner of the scene}."), whereas Pix4Cap only produces a incorrect description (e.g., ``A house appears in the woods at \textbf{bottom}."). 
In the last row, involving subtle changes and irrelevant variations, HiSem still successfully identifies fine-grained semantic changes, demonstrating robustness under challenging conditions. 
Overall, these qualitative captioning results demonstrate that our proposed method can accurately and comprehensively capture the fine-grained semantic information while suppressing irrelevant variations to enable more precise, detailed, and semantically faithful change captions.

\subsubsection{\textbf{Feature Visualization}}
\begin{figure}
	\centering
	\includegraphics[width=0.85\linewidth]{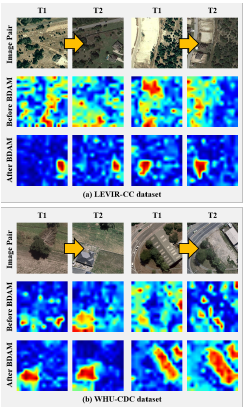}
	\caption{Visualization of features before and after BDAM module processing.
 }
	\label{fig:heatmap}
\end{figure}

To further validate the effectiveness of the BDAM module in extracting discriminative representations, we visualize feature maps for image pairs from the LEVIR-CC and WHU-CDC datasets before and after applying BDAM (Fig.~\ref{fig:heatmap}). For the LEVIR-CC examples (Fig.~\ref{fig:heatmap}(a)), which contain subtle changes, BDAM introduces discrepancy-guided cross-temporal interactions, producing consistent and accurate attention responses across both temporal observations. Similarly, for the WHU-CDC examples (Fig.~\ref{fig:heatmap}(b)), where the before-to-after transition involves both appearance and disappearance changes, the feature maps show that BDAM simultaneously emphasizes multiple types of true changes while suppressing irrelevant variations. These observations demonstrate that BDAM effectively guides directional cross-temporal interactions, yielding more discriminative representations that support fine-grained and robust change perception.

\subsection{Ablation Studies}
\subsubsection{\textbf{Effectiveness of BDAM and HASD}}
Table~\ref{tab:Ablation_HiSem_Core Modules} presents the ablation results of the proposed BDAM and HASD modules on the LEVIR-CC dataset. 
Starting from the baseline model, incorporating the BDAM module yields consistent improvements across all evaluation metrics, achieving gains of $+1.10\%$ on BLEU-4, $+5.63\%$ on CIDEr-D, and $+2.34\%$ on $S^*_m$. These results demonstrate that bidirectional discrepancy-guided attention effectively enhances discriminative bi-temporal representations and improves the localization of change regions. When HASD is further integrated, the model achieves the best performance across all metrics. Compared with the baseline, the full model improves BLEU-4 by $+1.50\%$, CIDEr-D by $+7.02\%$, and $S^*_m$ by $+3.05\%$. These results indicate that adaptive semantic routing and expert specialization effectively reduce semantic interference between changed and unchanged samples to obtain high-quality change-aware representations, enabling more accurate, detailed, and semantically consistent change descriptions.

\begin{table*} 
\renewcommand{\arraystretch}{1.3}
\caption{Ablation studies on the BDAM and HASD modules on the LEVIR-CC dataset. The bolded results are the best.
}
\label{tab:Ablation_HiSem_Core Modules}
\centering
\resizebox{0.95\linewidth}{!}{
\begin{tabular}{c c c | c c c c c c c | c}
	\toprule
	Method & BDAM & HASD & BLEU-1 & BLEU-2 & BLEU-3 & BLEU-4 & METEOR & ROUGE$_L$ & CIDEr-D & $S^*_m$\\
	\midrule
    {Baseline} &\ding{55} &\ding{55} & 84.51 & 76.73 &69.89 & 64.32 & 38.79 & 73.68 & 131.84 & 77.16 \\
    {--} &\ding{52} &\ding{55} & 86.07 & 78.16 & 71.17 & 65.42 & 39.94 & 75.14 & 137.47 & 79.50 \\
    {--} &\ding{55} &\ding{52} & 83.31 & 75.43 & 68.40 & 62.70 & 38.02 & 72.33 & 129.09 & 75.54 \\
 {HiSem (Ours)} &\ding{52} &\ding{52}
 & \textbf{86.60} & \textbf{78.74} & \textbf{71.73} & \textbf{65.82} & \textbf{40.39} & \textbf{75.77}	& \textbf{138.86} & \textbf{80.21} \\
	\bottomrule
\end{tabular}
}
\end{table*}

\begin{table*} [!t]
\renewcommand{\arraystretch}{1.3}
\caption{Ablation studies on the components of the BDAM module on the LEVIR-CC dataset, analyzing the effects of the Discrepancy-aware Feature Conditioning (DFC) sublayer and Bidirectional Discrepancy-guided Cross-temporal Interaction (BDCI) sublayer.
}
\label{tab:Ablation_BDAM}
\centering
\resizebox{0.95\linewidth}{!}{
\begin{tabular}{c c | c c c c c c c | c}
	\toprule
	DFC Sublayer & BDCI Sublayer & BLEU-1 & BLEU-2 & BLEU-3 & BLEU-4 & METEOR & ROUGE$_L$ & CIDEr-D & $S^*_m$\\
	\midrule
    \ding{55} &\ding{55} & 83.31 & 75.43 & 68.40 & 62.70 & 38.02 & 72.33 & 129.09 & 75.54 \\
    \ding{52} &\ding{55} & 83.73 & 75.57 & 68.44 & 62.65 & 38.25 & 72.49 & 130.13 & 75.88 \\
    \ding{55} &\ding{52} & 84.51 & 75.90 & 68.74 & 62.97 & 39.51 & 74.50 & 135.35 & 78.08 \\
 \ding{52} &\ding{52}
 & \textbf{86.60} & \textbf{78.74} & \textbf{71.73} & \textbf{65.82} & \textbf{40.39} & \textbf{75.77}	& \textbf{138.86} & \textbf{80.21} 
 \\
	\bottomrule
\end{tabular}
}
\end{table*}

\begin{table*}
\renewcommand{\arraystretch}{1.3}
\caption{Ablation studies on the components of the HASD module on the LEVIR-CC dataset, analyzing the effect of introducing coarse-grained image-level routing under different semantic modeling architectures (FFN vs. MoE).}
\label{tab:Ablation_HASD}
\centering
\begin{tabular}{c | cc cc | c c c c c c c | c}
\toprule
\multirow{2}{*}{\makecell[c]{Coarse-Grained \\Image-Level \\ Routing}}
& \multicolumn{2}{c}{Changed Path} 
& \multicolumn{2}{c}{Unchanged Path}
& \multirow{2}{*}{BLEU-1} 
& \multirow{2}{*}{BLEU-2} 
& \multirow{2}{*}{BLEU-3} 
& \multirow{2}{*}{BLEU-4} 
& \multirow{2}{*}{METEOR} 
& \multirow{2}{*}{ROUGE$_L$} 
& \multirow{2}{*}{CIDEr-D}
& \multirow{2}{*}{$S^*_m$} \\
\cmidrule(lr){2-3} \cmidrule(lr){4-5}
& FFN & MoE & FFN & MoE &  &  &  &  &  &  &  & \\
\midrule
\ding{55} & \ding{55} & \ding{55} & \ding{55} & \ding{55} & 86.07 & 78.16 & 71.17 & 65.42 & 39.94 & 75.14 & 137.47 & 79.50 \\
\ding{55} & \ding{52} & \ding{55} & \ding{52} & \ding{55} & 86.17 & 78.20 & 71.02 & 64.99 & 40.00 & 75.18 & 137.00 & 79.29 \\
\ding{55} & \ding{55} & \ding{52} & \ding{55} & \ding{52} & 86.38 & 78.45 & 71.34 & 65.38 & 40.37 & 75.70 & 138.65 & 80.02 \\
\ding{52} & \ding{52} & \ding{55} & \ding{52} & \ding{55} & 85.42 & 77.59 & 70.75 & 65.15 & 39.90 & 75.00 & 136.42 & 79.12 \\
\ding{52} & \ding{55} & \ding{52} & \ding{55} & \ding{52} & 85.72 & 77.90 & 70.87 & 65.10 & 39.62 & 75.02 & 135.87 & 78.90 \\
\ding{52} & \ding{55} & \ding{52} & \ding{52} & \ding{55}
& \textbf{86.60} & \textbf{78.74} & \textbf{71.73} & \textbf{65.82} & \textbf{40.39} & \textbf{75.77}	& \textbf{138.86} & \textbf{80.21} \\
\bottomrule
\end{tabular}
\end{table*}

\subsubsection{\textbf{Detailed Component Effectiveness Analysis of BDAM}}
Table~\ref{tab:Ablation_BDAM} investigates the contribution of each sublayer within the BDAM module, containing the Discrepancy-aware Feature Conditioning (DFC) and the Bidirectional Discrepancy-guided Cross-temporal Interaction (BDCI). The results reveal a compelling synergistic effect between two sublayers to facilitate the modeling of synergistic spatial and temporal information.
Specifically, while the independent integration of the DFC sublayer yields relatively stable performance, it serves as a critical representation foundation by refining the spatial representation of bi-temporal features. The modest gains observed with DFC alone suggest that purely spatial enhancement, without effective temporal reasoning, is insufficient for complex change recognition. On the other hand, the BDCI sublayer provides a more significant performance boost (e.g., $+6.26\%$ in CIDEr-D) by utilizing bidirectional discrepancies as structural priors to guide cross-temporal attention. This validates the importance of explicit discrepancy modeling in capturing authentic changes.

The most crucial observation is the significant performance leap achieved through the joint integration of both sublayers. The full BDAM configuration significantly outperforms the baseline by $+3.12\%$ in BLEU-4 and a substantial $+9.77\%$ in CIDEr-D, which exceeds the cumulative gains of the two individual sublayers. This significant improvement demonstrates that DFC effectively ``recalibrates'' the feature space, allowing the subsequent BDCI to more precisely extract semantic changes. These results strongly validate our design philosophy: the sequential transition from spatial feature conditioning to discrepancy-guided temporal interaction is not merely additive but fundamentally essential for decoding the intricate dynamics of remote sensing imagery.

\subsubsection{\textbf{Effectiveness of Hierarchical Semantic Disentanglement}}
Table~\ref{tab:Ablation_HASD} evaluates the core design philosophy of the HASD module: the decoupling of bi-temporal change perception into a hierarchical routing and task-tailored modeling process. 
In a monolithic architecture (Rows 2-3 in Table~\ref{tab:Ablation_HASD}) where no image-level routing is applied, replacing standard FFNs with MoE consistently yields performance gains across all metrics (e.g., $+1.65\%$ in CIDEr-D). This confirms that MoE's sparse activation mechanism possesses a superior inductive bias for capturing the heterogeneous semantic changes inherent in remote sensing temporal imagery compared to static FFNs.
A pivotal observation arises when introducing the coarse-grained image-level routing. Interestingly, maintaining a \textit{symmetric} architecture for both paths (Rows 4-5) leads to a performance bottleneck or even degradation compared to their monolithic counterparts. This phenomenon suggests that once the router explicitly disentangles the ``changed'' and ``unchanged'' semantic streams, employing identical representational capacity for both becomes sub-optimal. Unchanged samples, characterized by high semantic homogeneity, do not require the high-dimensional expert space of an MoE; forcing them through a complex routing-based ensemble may introduce unnecessary noise or over-fitting.

The optimal performance is achieved in Row 6, where we implement an \textit{asymmetric strategy}: a specialized MoE for the high-entropy ``changed'' path and a parsimonious FFN for the low-entropy ``unchanged'' path. This configuration reaches the state-of-the-art across all benchmarks, notably achieving $138.86$ in CIDEr-D and $80.21$ in $S^*_m$. These results empirically validate our hypothesis: the RSICC task is best addressed by first explicitly bifurcating the data stream and then aligning the model capacity with the intrinsic semantic complexity of each sub-task. This ``divide-and-conquer'' approach not only enhances descriptive precision for complex changes but also ensures structural efficiency for consistent scenes.

\begin{table*}
\renewcommand{\arraystretch}{1.3}
\caption{Performance analysis of our semantic disentanglement mechanism on the LEVIR-CC dataset. We compare two inference settings: \textit{Routing (w. Pre)}, which relies on predicted change existence, and \textit{Routing (w. GT)}, which employs ground-truth labels to eliminate routing uncertainty. Notably, we introduce a normalized performance gain defined as $\rho$, which measures the conversion efficiency from routing improvements to language generation gains.}
\label{tab:Study_Advantages_Paradigm_LEVIR-CC}
\centering
\begin{tabular}{l|l|c|cccccc|c|c}
\toprule
\multicolumn{1}{c|}{Evaluation Scope} & \multicolumn{1}{c|}{Method} & $Acc_\text{router}$ & BLEU-1 & BLEU-2 & BLEU-3 & BLEU-4 & METEOR & ROUGE-L & CIDEr-D & $S_m^*$ \\
\midrule
\multirow{4}{*}{\makecell[l]{Unchanged\\image pairs}} 
& Routing \textit{(w. Pre.)} & 96.91\% & 97.22 & 96.80 & 96.57 & 96.41 & 75.75 & 97.55 & -- & -- \\
& Routing \textit{(w. GT)} & 100\% & 100.00 & 100.00 & 100.00 & 100.00 & 100.00 & 100.00 & -- & -- \\
& Improvement ($\Delta$) & \textbf{+3.09\%} & \textbf{+2.78} & \textbf{+3.20} & \textbf{+3.43} & \textbf{+3.59} & \textbf{+24.25} & \textbf{+2.45} & -- & -- \\
& \multicolumn{1}{l|}{\cellcolor{gray!15}$\rho = \Delta Score\ / \Delta Acc_{\text{router}}$} & \cellcolor{gray!15}\textbf{--} & \cellcolor{gray!15}\textbf{0.90} & \cellcolor{gray!15}\textbf{1.04} & \cellcolor{gray!15}\textbf{1.11} & \cellcolor{gray!15}\textbf{1.16} & \cellcolor{gray!15}\textbf{7.85} & \cellcolor{gray!15}\textbf{0.79} & \cellcolor{gray!15}\textbf{--} & \cellcolor{gray!15}\textbf{--}  \\
\midrule

\multirow{4}{*}{\makecell[l]{Changed\\image pairs}} 
& Routing \textit{(w. Pre)} & 91.06\% & 77.89 & 63.94 & 51.15 & 40.41 & 25.92 & 53.96 & 65.12 & 46.35 \\
& Routing \textit{(w. GT)} & 100\% & 79.69 & 65.96 & 52.95 & 41.78 & 27.05 & 56.32 & 69.54 & 48.67 \\
& Improvement ($\Delta$) & \textbf{+8.94\%} & \textbf{+1.80} & \textbf{+2.02} & \textbf{+1.80} & \textbf{+1.37} & \textbf{+1.13} & \textbf{+2.36} & \textbf{+4.42} & \textbf{+2.32} \\
& \multicolumn{1}{l|}{\cellcolor{gray!15}$\rho = \Delta Score\ / \Delta Acc_{\text{router}}$} & \cellcolor{gray!15}\textbf{--} & \cellcolor{gray!15}\textbf{0.20} & \cellcolor{gray!15}\textbf{0.23} & \cellcolor{gray!15}\textbf{0.20} & \cellcolor{gray!15}\textbf{0.15} & \cellcolor{gray!15}\textbf{0.13} & \cellcolor{gray!15}\textbf{0.26} & \cellcolor{gray!15}\textbf{0.49} & \cellcolor{gray!15}\textbf{0.26}  \\
\midrule

\multirow{4}{*}{\makecell[l]{All image\\pairs}} 
& Routing \textit{(w. Pre)} & 93.99\% & 86.60 & 78.74 & 71.73 & 65.82 & 40.39 & 75.77 & 138.86 & 80.21 \\
& Routing \textit{(w. GT)} & 100\% & 88.53 & 80.91 & 73.82 & 67.69 & 41.89 & 78.17 & 144.63 & 83.09 \\
& Improvement ($\Delta$) & \textbf{+6.01\%} & \textbf{+1.93} & \textbf{+2.17} & \textbf{+2.09} & \textbf{+1.87} & \textbf{+1.50} & \textbf{+2.40} & \textbf{+5.77} & \textbf{+2.88} \\
& \multicolumn{1}{l|}{\cellcolor{gray!15}$\rho = \Delta Score\ / \Delta Acc_{\text{router}}$} & \cellcolor{gray!15}\textbf{--} & \cellcolor{gray!15}\textbf{0.32} & \cellcolor{gray!15}\textbf{0.36} & \cellcolor{gray!15}\textbf{0.35} & \cellcolor{gray!15}\textbf{0.31} & \cellcolor{gray!15}\textbf{0.25} & \cellcolor{gray!15}\textbf{0.40} & \cellcolor{gray!15}\textbf{0.96} & \cellcolor{gray!15}\textbf{0.48}  \\
\bottomrule
\end{tabular}
\end{table*}

\begin{table*}
\renewcommand{\arraystretch}{1.3}
\caption{Performance analysis of our semantic disentanglement mechanism on the WHU-CDC dataset. We compare two inference settings: \textit{Routing (w. Pre)}, which relies on predicted change existence, and \textit{Routing (w. GT)}, which employs ground-truth labels to eliminate routing uncertainty. Notably, we introduce a normalized performance gain defined as $\rho$, which measures the conversion efficiency from routing improvements to language generation gains.}
\label{tab:Study_Advantages_Paradigm_WHU-CDC}
\centering
\begin{tabular}{l|l|c|cccccc|c|c}
\toprule
\multicolumn{1}{c|}{Evaluation Scope} & \multicolumn{1}{c|}{Method} & $Acc_\text{router}$ & BLEU-1 & BLEU-2 & BLEU-3 & BLEU-4 & METEOR & ROUGE-L & CIDEr-D & $S_m^*$ \\
\midrule
\multirow{4}{*}{\makecell[l]{Unchanged\\image pairs}} 
& Routing \textit{(w. Pre)} & 97.40\% & 98.07 & 97.85 & 97.77 & 97.74 & 77.94 & 97.89 & -- & -- \\
& Routing \textit{(w. GT)} & 100\% & 100.00 & 100.00 & 100.00 & 100.00 & 100.00 & 100.00 & -- & -- \\
& Improvement ($\Delta$) & \textbf{+2.60\%} & \textbf{+1.93} & \textbf{+2.15} & \textbf{+2.23} & \textbf{+2.26} & \textbf{+22.06} & \textbf{+2.11} & -- & -- \\
& \multicolumn{1}{l|}{\cellcolor{gray!15}$\rho = \Delta Score\ / \Delta Acc_{\text{router}}$} & \cellcolor{gray!15}\textbf{--} & \cellcolor{gray!15}\textbf{0.74} & \cellcolor{gray!15}\textbf{0.83} & \cellcolor{gray!15}\textbf{0.86} & \cellcolor{gray!15}\textbf{0.87} & \cellcolor{gray!15}\textbf{8.48} & \cellcolor{gray!15}\textbf{0.81} & \cellcolor{gray!15}\textbf{--} & \cellcolor{gray!15}\textbf{--}  \\
\midrule

\multirow{4}{*}{\makecell[l]{Changed\\image pairs}} 
& Routing \textit{(w. Pre)} & 80.08\% & 65.51 & 50.92 & 37.89 & 28.49 & 22.45 & 47.78 & 58.98 & 39.42 \\
& Routing \textit{(w. GT)} & 100\% & 70.97 & 57.42 & 43.80 & 33.15 & 25.12 & 53.66 & 71.49 & 45.86 \\
& Improvement ($\Delta$) & \textbf{+19.92\%} & \textbf{+5.46} & \textbf{+6.50} & \textbf{+5.91} & \textbf{+4.66} & \textbf{+2.67} & \textbf{+5.88} & \textbf{+12.51} & \textbf{+6.44} \\
& \multicolumn{1}{l|}{\cellcolor{gray!15}$\rho = \Delta Score\ / \Delta Acc_{\text{router}}$} & \cellcolor{gray!15}\textbf{--} & \cellcolor{gray!15}\textbf{0.27} & \cellcolor{gray!15}\textbf{0.33} & \cellcolor{gray!15}\textbf{0.30} & \cellcolor{gray!15}\textbf{0.23} & \cellcolor{gray!15}\textbf{0.13} & \cellcolor{gray!15}\textbf{0.30} & \cellcolor{gray!15}\textbf{0.63} & \cellcolor{gray!15}\textbf{0.32}  \\
\midrule

\multirow{4}{*}{\makecell[l]{All image\\pairs}} 
& Routing \textit{(w. Pre)} & 91.91\% & 87.49 & 82.89 & 79.17 & 76.52 & 48.77 & 82.00 & 158.35 & 91.41 \\
& Routing \textit{(w. GT)} & 100\% & 90.37 & 86.16 & 82.26 & 79.28 & 51.06 & 85.30 & 167.27 & 95.73 \\
& Improvement ($\Delta$) & \textbf{+8.09\%} & \textbf{+2.88} & \textbf{+3.27} & \textbf{+3.09} & \textbf{+2.76} & \textbf{+2.29} & \textbf{+3.30} & \textbf{+8.92} & \textbf{+4.32} \\
& \multicolumn{1}{l|}{\cellcolor{gray!15}$\rho = \Delta Score\ / \Delta Acc_{\text{router}}$} & \cellcolor{gray!15}\textbf{--} & \cellcolor{gray!15}\textbf{0.36} & \cellcolor{gray!15}\textbf{0.40} & \cellcolor{gray!15}\textbf{0.38} & \cellcolor{gray!15}\textbf{0.34} & \cellcolor{gray!15}\textbf{0.28} & \cellcolor{gray!15}\textbf{0.41} & \cellcolor{gray!15}\textbf{1.10} & \cellcolor{gray!15}\textbf{0.53}  \\
\bottomrule
\end{tabular}
\end{table*}

\begin{table*} [!t]
\renewcommand{\arraystretch}{1.3}
\caption{performance of the model in different depths on the LEVIR-CC dataset. L.B. denotes the depth of the BDAM, and L.d. denotes the depth of the caption decoder.
}
\label{tab:Ablation_Depth}
\centering
\resizebox{0.8\linewidth}{!}{
\begin{tabular}{c c | c c c c c c c | c}
	\toprule
	L.B. & L.D. & BLEU-1 & BLEU-2 & BLEU-3 & BLEU-4 & METEOR & ROUGE$_L$ & CIDEr-D & $S^*_m$\\
	\midrule
    1 & 1 & 85.99 & 77.93 & 70.72 & 64.74 & 40.24 & 75.42 & 138.11 & 79.63 \\
    2 & 1 & 86.11 & 78.13 & 71.14 & 65.33 & 40.13 & 75.24 & 137.55 & 79.56 \\
    3 & 1
 & \textbf{86.60} & \textbf{78.74} & \textbf{71.73} & \textbf{65.82} & \textbf{40.39} & \textbf{75.77} & \textbf{138.86} & \textbf{80.21} \\
    3 & 2 & 85.94 & 78.01 & 70.86 & 64.91 & 39.72 & 74.79 & 136.31 & 78.93 \\
    3 & 3 & 86.07 & 77.96 & 70.94 & 65.19 & 40.24 & 75.46 & 137.64 & 79.63 \\
	\bottomrule
\end{tabular}
}
\end{table*}

\subsection{Insights into Hierarchical Semantic Disentanglement}

The proposed hierarchical semantic disentanglement mechanism provides a structured perspective for RSICC by decomposing the task into two sequential stages: coarse-grained change existence perception and fine-grained change semantic understanding. By explicitly separating changed and unchanged image pairs, this design alleviates semantic entanglement and enables a divide-and-conquer modeling strategy, where different components can be designed for distinct semantic objectives.

To further analyze the implications of this strategy, we conduct additional experiments on the LEVIR-CC and WHU-CDC datasets (Tables~\ref{tab:Study_Advantages_Paradigm_LEVIR-CC} and \ref{tab:Study_Advantages_Paradigm_WHU-CDC}). We compare two inference settings: \textit{Routing (w. Pre)}, which relies on predicted change existence, and \textit{Routing (w. GT)}, which employs ground-truth labels to eliminate routing uncertainty, thereby representing the upper-bound performance of the proposed framework. Notably, to quantify how routing accuracy translates into caption quality, we introduce a normalized performance gain defined as $\rho = \Delta\text{Score} / \Delta Acc_{\text{router}}$, which measures the conversion efficiency from routing improvements to language generation gains.
In addition, we omit the CIDEr-D scores for unchanged image pairs in Tables VI and VII because the generated captions are highly repetitive, leading to score collapse (i.e., near-zero values) and negligible variance across samples. As a result, CIDEr-D becomes uninformative for comparison in this setting.

\textbf{Justification of disentangling change existence and semantic understanding.}
A consistent observation across both datasets is that, for changed image pairs, substantial improvements in routing accuracy under ground-truth guidance (+8.94\% on LEVIR-CC and +19.92\% on WHU-CDC) lead to only moderate gains in captioning performance (+1.37\% and +4.66\% on BLEU-4, respectively), resulting in significantly lower normalized gains compared to unchanged samples. 
This phenomenon indicates that, while coarse-grained change existence perception can be improved relatively reliably, capturing the underlying fine-grained semantics of changes remains considerably more challenging. Such a discrepancy reveals an intrinsic property of RSICC: changed and unchanged image pairs exhibit fundamentally different levels of semantic complexity. Unchanged pairs primarily require coarse-grained semantic confirmation, whereas changed pairs demand detailed modeling of object attributes, spatial relationships, and temporal evolution. This inherent asymmetry justifies the necessity of explicitly disentangling their representations, as uniform modeling inevitably leads to semantic interference and suboptimal performance.

\textbf{Routing accuracy as a bottleneck for coarse-to-fine modeling.}
Across both datasets, we observe a consistent positive correlation between routing accuracy and captioning performance, suggesting that the quality of the routing signal plays a pivotal role in the overall framework. This finding identifies coarse-grained change existence perception as a critical bottleneck, emphasizing the importance of enhancing the discriminative capability of the BDAM module to provide reliable inputs for subsequent fine-grained semantic modeling.

\textbf{From coarse-grained routing to fine-grained semantic modeling.}
Despite its effectiveness, the relatively low conversion efficiency observed for changed samples indicates that accurately identifying change existence does not directly translate into equally strong improvements in fine-grained semantic description. In our framework, this gap is explicitly addressed by the HASD module, where a MoE-based design enables token-level modeling of diverse change patterns by assigning different experts to specialize in distinct semantic variations, thereby moving beyond binary routing toward fine-grained semantic understanding. Nevertheless, the remaining discrepancy suggests that modeling complex real-world changes remains challenging, and further improvements may be achieved by enhancing expert specialization and designing more expressive routing strategies to better align coarse-grained perception with high-fidelity semantic generation.

\textbf{Implications for future RSICC research.}
Overall, the proposed mechanism introduces a clear separation between change existence perception and semantic understanding, providing a structured guideline for RSICC modeling. This divide-and-conquer strategy enables the design of specialized components for different semantic granularities, thereby reducing semantic confusion and improving descriptive accuracy in complex remote sensing scenarios.

\subsection{{Parametric Experiments and Analysis}}

\textbf{{Layer Depth Analysis of BDAM and Caption Decoder:}}
To further investigate the effect of model depth, we conduct an ablation study on both the BDAM module and the Transformer-based caption decoder, as shown in Table~\ref{tab:Ablation_Depth}. L.B. denotes the depth of the BDAM, and L.D. denotes the depth of the caption decoder. The results indicate that the optimal configuration is a 3-layer BDAM with a 1-layer decoder. This suggests that a deeper BDAM helps learn more discriminative change features, whereas increasing the decoder depth does not further improve captioning performance. Accordingly, we use L.B.=3 and L.D.=1 in the final model.

\section{Conclusion}
In this paper, we reveal a fundamental limitation in existing RSICC methods: the unified modeling of changed and unchanged image pairs overlooks their inherently different semantic complexities.
To address this issue, we propose a hierarchical semantic disentanglement network (HiSem), which explicitly decomposes the task into coarse-grained change existence perception and fine-grained change semantic understanding. 
Building upon this insight, the proposed BDAM module enhances discriminative change existence perception via discrepancy-aware cross-temporal interaction, while the HASD module enables adaptive and specialized modeling through hierarchical routing and expert-based semantic understanding. 
Extensive experiments demonstrate that HiSem consistently achieves superior performance across two benchmarks. More importantly, this work provides a structured perspective for RSICC by explicitly aligning model design with the intrinsic semantic heterogeneity of bi-temporal scenes.

\ifCLASSOPTIONcaptionsoff
\newpage
\fi

\bibliographystyle{IEEEtran}
\bibliography{papers.bib}

\end{document}